\documentclass{article}

\usepackage{arxiv}

\usepackage[utf8]{inputenc} % allow utf-8 input
\usepackage[T1]{fontenc}    % use 8-bit T1 fonts
\usepackage{hyperref}       % hyperlinks
\usepackage{url}            % simple URL typesetting
\usepackage{booktabs}       % professional-quality tables
\usepackage{amsfonts}       % blackboard math symbols
\usepackage{nicefrac}       % compact symbols for 1/2, etc.
\usepackage{microtype}      % microtypography
\usepackage{lipsum}		% Can be removed after putting your text content
\usepackage{graphicx}
\usepackage{natbib}
\usepackage{doi}

\usepackage{amsmath,graphicx}
\usepackage{textcomp}
\usepackage{siunitx} 
\usepackage{array}   
\usepackage{multirow}
\usepackage{tabularx}

\usepackage{enumitem}

\title{Adaptive Smooth Activation for Improved Disease Diagnosis and Organ Segmentation from Radiology Scans}

\date{} 					% Or removing it

\author{Koushik Biswas, Debesh Jha, Nikhil Kumar Tomar, Gorkem Durak, Alpay Medetalibeyoglu,\\ \textbf{Matthew Antalek,  Yury Velichko, Daniela Ladner, Amir Bohrani,   Ulas Bagci}}

\date{Machine \& Hybrid Intelligence Lab, Department of Radiology, Northwestern University}

\renewcommand{\shorttitle}{Adaptive Smooth Activation for Improved Disease Diagnosis}

\hypersetup{
pdftitle={Smooth Activation Function for IMPROVED DISEASE
DIAGNOSIS AND ORGAN SEGMENTATION},
pdfsubject={},
pdfauthor={Koushik Biswas, Debesh Jha, Nikhil Kumar Tomar, Gorkem Durak, Alpay Medetalibeyoglu, Matthew Antalek,  Yury Velichko, Daniela Ladner, Amir Bohrani,   Ulas Bagci},
pdfkeywords={Deep Learning, Medical Imaging},
}

\begin{document}
\maketitle

\begin{abstract}
In this study, we propose a new activation function, called Adaptive Smooth Activation Unit (ASAU), tailored for optimized gradient propagation, thereby enhancing the proficiency of convolutional networks in medical image analysis. We apply this new activation function to two important and commonly used general tasks in medical image analysis: automatic disease diagnosis and organ segmentation in CT and MRI. %Accurate abdominal organ classification (from CT and MRI) and precise liver segmentation are essential for computer-aided diagnosis, surgical planning, and treatment monitoring. Conventional activation functions within deep learning frameworks often struggle with anatomical variability and intensity inhomogeneities. To address this, we introduce the Adaptive Smooth Activation Unit (ASAU), a robust activation function designed for seamless gradient flow, ensuring improved performance of deep learning models in medical imaging tasks. 
Our rigorous evaluation on the RadImageNet abdominal/pelvis (CT and MRI) dataset and Liver Tumor Segmentation Benchmark (LiTS) 2017 demonstrates that our ASAU-integrated frameworks not only achieve a  substantial (4.80\%) improvement over ReLU in classification accuracy (disease detection) on abdominal CT and MRI  but also achieves 1\%-3\% improvement in dice coefficient compared to widely used activations for `healthy liver tissue' segmentation. These improvements offer new baselines for developing a diagnostic tool, particularly for complex, challenging pathologies. The superior performance and adaptability of ASAU highlight its potential for integration into a wide range of image classification and segmentation tasks.
\end{abstract}

% keywords can be removed

\section{Introduction}
\label{sec:introduction}
\textbf{Computer Aided Detection (Classification):} Detecting pathologies from CT and MRI scans automatically is crucial for accurate diagnosis, early detection, and treatment planning. Abdominal CT exams are cost-effective and widely available, making them essential for emergency conditions and detecting lesions. They provide quick and high-resolution images that are particularly useful for identifying acute abdominal pathologies and abdominal tumors. On the other hand, abdominal MRI images provide superior soft tissue contrast and are more effective in classifying tumor types without using ionizing radiation. 
Automated disease detection from abdominal scans (CT, MRI) plays a pivotal role in the rapid and accurate diagnosis of critical emergencies such as acute pancreatitis, acute appendicitis, and various tumoral lesions affecting organs such as the pancreas, liver, and kidneys. Missed diagnoses of conditions like acute pancreatitis can be life-threatening, which highlights the critical role of automated detection systems in preventing potential mortalities.

Early detection is particularly important for conditions like precursor pancreatic lesions, where timely identification can significantly impact prognosis and patient outcomes. The potential occurrence of pancreatic cancer underlines the importance of rapid intervention at an earlier stage, given the poor prognosis associated with delayed detection. The automatic detection and classification of liver lesions as benign or malignant is crucial for follow-up procedures and treatment strategies. Similarly, detecting lesions in organs like the prostate and adrenal glands helps in patient management and prognosis. Accurately interpreting abdominal scans for these conditions is essential for guiding healthcare professionals in making informed decisions about treatment approaches and ensuring optimal patient care. After all, automated disease detection in abdominal imaging not only improves diagnostic accuracy but also has the potential to significantly enhance prognosis, patient management, and the quality of healthcare services.

Abdominal anatomy is complex and variable, with numerous organs and structures closely packed together. As a result, distinguishing between healthy and pathological tissues in earlier phases can be challenging due to subtle differences~\cite{reyes2008interpretability}. This complexity can lead to variability in interpretation among radiologists, particularly with rare or special conditions. CAD-based algorithms can be helpful in reducing diagnostic errors and improving diagnosis.

%The complexity and variability of abdominal anatomy, where many organs and structures are closely packed together, often pose subtle differences between healthy and pathological tissues in earlier phases. This complexity can lead to variability in interpretation among radiologists, especially with rare or nuanced pathologies. Therefore, CAD-based algorithms can help reduce diagnostic errors and improve diagnosis. 

%It will help the diagnostic process be more efficient for the radiologists, leading to better patient care and resource management. 

\textbf{Organ Segmentation:} For segmentation motivation, we would like to focus on the liver organ for an example, although what we propose herein is generic and can be applied to any organ, pathology, and non-medical applications. We choose the liver because liver cancer is the third most common cause of cancer-related death worldwide~\cite{arnold2020global}, and the liver needs to be quantified either by volume for several diseases, and also its labelled boundary is used for CAD systems where algorithms seek for lesions inside the liver volume. Besides, the liver is one of the most challenging organs to segment because of its highly variable shape and close proximity to other organs. %Manual liver segmentation is challenging because it is time-consuming and lacks reproducibility~\cite {gotra2017liver}. 
Liver segmentation is a prerequisite also for the localization of tumors for targeted therapies and the detailed planning of surgical interventions. It demands the highest levels of precision, as the outcomes can directly influence therapeutic decisions and patient prognosis. The variability in contrast, shape, and appearance of liver tissue poses significant challenges. Therefore, automated methods can play an essential role in handling such complexity.

\textbf{Domain specific activation functions are needed:} Current deep learning approaches have shown promise in enhancing the quality of classification and segmentation tasks~\cite{jha2020doubleu,tomar2022transresu,jha2023transnetr}. However, they are often constrained by the limitations of the activation functions utilized, particularly in convolutional neural networks (CNNs). Traditional activation functions such as ReLU~\cite{relu} and its variant (for example, Leaky ReLU~\cite{xu2015empirical}, PReLU~\cite{he2015delving}), despite being instrumental to the success of CNN, exhibit notable shortcomings. They are prone to information loss in areas of negative input and often fail to capture the subtle nuances necessary for delineating intricate anatomical features, leading to potential inaccuracies in segmentations, which are highly undesirable in a clinical setting. This necessitates the development of more specialized functions tailored to the domain-specific needs of medical image analysis.

\textbf{Our proposal:} Here, we present a novel smooth activation function called the Adaptive Smooth Activation Function (ASAU), carefully designed to handle the challenges prevalent in medical image analysis. This function embodies a methodological shift towards smoother, more continuous transitions, offering refined gradients that promote the intricate learning necessary for high-fidelity classification and segmentation tasks. With the rigorous testing and empirical evaluation on classification and segmentation tasks, our research indicates that applying this ASAU within conventional CNN architectures can substantially enhance the classification for multiclass disease detection from abdominal scans and liver segmentation from CT scans. Our main contributions are:

\begin{itemize}
    \item We have introduced a novel Adaptive Smooth Activation Function (ASAU) that significantly enhances the classification (disease detection) and segmentation (liver) accuracy. 
    
    \item Our extensive comparative analysis using Abdomen/Pelvis benchmark classification and healthy liver segmentation datasets in both CT and MRI modalities demonstrates almost 5\% improvement in accuracy in both CT and MRI cases and 1\% improvement for segmentation tasks (for an organ such as liver, 1\% is a significant increment) that show the effectiveness of ASAU.

%\item To our best of knowledge, this is the first study showing an effectiveness of a domain-tuned activation function applied to both classification and segmentation tasks at the same time, with rigorous evaluation and significantly improved results confirming new state of the art.

    %\item We obtained an improvement of 6\% for the classification tasks and 1\% for the segmentation tasks, demonstrating the effectiveness of the proposed method. 
    
\end{itemize}

% \begin{itemize}[noitemsep] 
%   \item \textbf{Smooth Activation Function:} We introduce a novel smooth activation function within the SmoothDoubleUNet, which mitigates the effects of noise and enhances boundary delineation in segmentation tasks, leading to higher precision.
  
%   \item \textbf{Training with Variable Perturbations:} The model incorporates a robust training regimen that uses variable perturbations, boosting the network's ability to handle input variability and significantly improving its generalization across diverse datasets.
  
%   \item \textbf{Selective Encoder Fine-tuning:} Our selective fine-tuning approach, which focuses on the image encoders while maintaining fixed mask decoders, optimizes feature extraction and demonstrates the practical efficacy of transfer learning in medical image segmentation.
% \end{itemize}

%%%%%%%%%%%%%%%%%%%%%%%%%%%%%%%%%%%%%%%%%%%%%%%%%%%%%%%%%
\begin{figure*}[!t]
\begin{minipage}[t]{.32\linewidth}
        \centering
         \includegraphics[width=\linewidth]{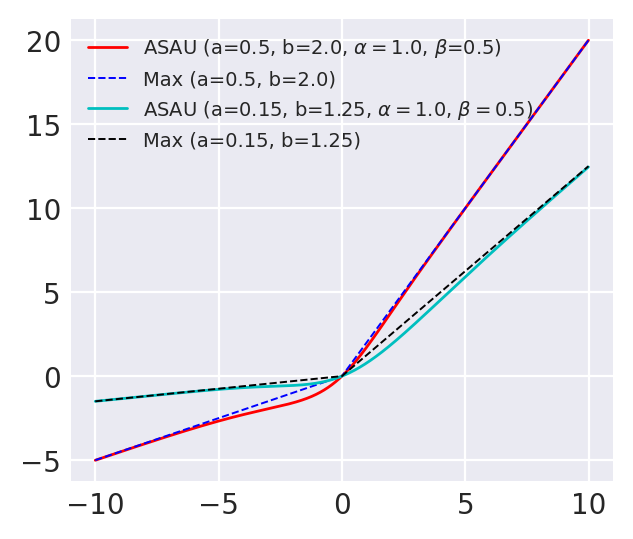}
        
        \caption{Approximation of Maximum function using ASAU for varying values of a, b, $\alpha$, and $\beta$. As $\beta \rightarrow \infty$, ASAU smoothly approximates the maximum function.}
        \label{ASAU}
         \end{minipage}
         \hfill
   \begin{minipage}[t]{.32\linewidth}
        \centering
    
         \includegraphics[width=\linewidth]{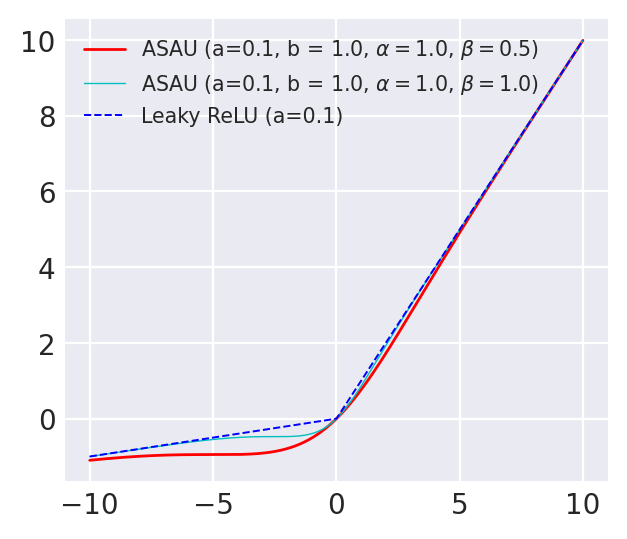}
        
        \caption{Approximation of Leaky ReLU using ASAU for different values of a, b, $\alpha$, and $\beta$. As $\beta \rightarrow \infty$, ASAU smoothly approximates Leaky ReLU.}
        \label{ASAU1}
   \end{minipage}
    \hfill
   \begin{minipage}[t]{.32\linewidth}
        \centering
    
         \includegraphics[width=\linewidth]{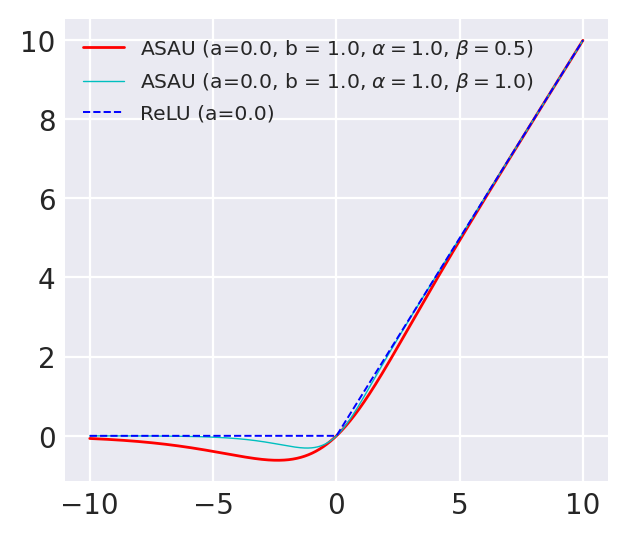}
       
        \caption{Approximation of ReLU using ASAU for different values of a, b, $\alpha$, and $\beta$. As $\beta \rightarrow \infty$, ASAU smoothly approximates ReLU.}
        \label{ASAU2}
   \end{minipage}
   \vspace{-5mm}
\end{figure*}

%%%%%%%%%%%%%%%%%%%%%%%%%%%%%%%%%%%%%%%%%%%%%%%%%%

%%%%%%%%%%%%%%%%%%%%%%%%%%%%%%%%%%%%%%%%%%%%%%%%%%%%%%%%%%%%%%%%%%%%
\section{Method}
\label{sec:method}
\subsection{Adaptive Smooth Activation Function} 
We present the Adaptive Smoothing Activation Unit (ASAU) as a way of learning to activate the neurons or not. In this paper, we show how to approximate the maximum function using a smooth function. We consider the maximum function of two values, $\text{max}(x_{1}, x_{2})$, which is defined as follows: 
%---
\begin{equation}
\text{max}(x_{1}, x_{2}) =   
\begin{cases}
x_{1},& \text{if } x_{1} \geq x_{2} \\
x_{2},              & \text{otherwise}
\end{cases}
\label{Eq:max_func}
\end{equation}
%---
We can rewrite the equation~(\ref{Eq:max_func}) as follows:
%---
\begin{equation}
\text{max} (x_{1}, x_{2}) = x_{1} + \text{max}(0, x_{2} -  x_{1}). 
\label{Eq:max_func2}
\end{equation}

Note that the maximum function is not differentiable at the origin, but we need differentiable functions during backpropagation. Also, note that the popular activation functions like ReLU \cite{relu}, Leaky ReLU \cite{lrelu}, and Parametric ReLU are special cases of the maximum function, and they are not differentiable at the origin. Mish \cite{mish} is a smooth activation function that is handcrafted and fixed. If we can add a parameter $\beta$ in Mish, it can approximate the ReLU activation function smoothly. Note that, 
\begin{equation}\label{eq3}
    \text{if}\ \beta \rightarrow \infty,\\ \text{then}\ xtanh(\alpha \text{SoftPlus}(\beta x))\approx max(0,x) 
\end{equation}
where $\alpha$ is another smoothing parameter and SoftPlus is defined as $ln(1+e^x)$. Maxout activation \cite{maxout} is a linear combination of the convex functions that generalize the ReLU or its variants. Replacing the equation (\ref{eq3}) in equation (\ref{Eq:max_func2}), we have
\begin{equation}
    \text{max} (x_{1}, x_{2}) \approx x_{1} + (x_2-x_1)tanh(\alpha \text{SoftPlus}(\beta (x_2-x_1))
\end{equation}
If we consider $x_1 = ax$ and $x_2 = bx$ we can get an approximation of the Maxout family; in particular, we can derive the smooth approximation of the Leaky ReLU and ReLU activation function (by considering $a=1$ or $a=0$ respectively). 
\begin{equation}
    \text{max} (ax, bx) \approx ax + (b-a)xtanh(\alpha \text{SoftPlus}(\beta (b-a)x)
\end{equation}
Figure~(\ref{ASAU}) shows how the maxout has been approximated with the ASAU function. Similarly, Figure~(\ref{ASAU1}) and Figure~(\ref{ASAU2}) show how the Leaky ReLU and ReLU are approximated with the ASAU function, respectively. 
%%%%%%%%%%%%%%%%%%%%%%%%%%%%%%%%%%%%%%%%%%%%%%%%%%%%%%%%%%%%%%%%%%%%%

\section{Experiments}
\label{sec:experiments}

\begin{table*}[!t]
\scriptsize
    \centering
    \caption{Activation functions and their impact on RadImageNet 28 classes Abdominal/Pelvis CT Scans.}
    \begin{tabular}{c|c|c|c|c|c|c|c|c|c}
       \toprule
        \multirow{2}{*}{\shortstack{\textbf{Activation}\\ \textbf{Function}}} & \multirow{2}{*}{\textbf{Method}} & \multicolumn{3}{c|}{\textbf{Macro Average}} & \multicolumn{3}{c|}{\textbf{Micro Average}} & \multirow{2}{*}{\textbf{Accuracy}}  & \multirow{2}{*}{\textbf{MCC}} \\
        \cline{3-8}
        & & \textbf{Precision} & \textbf{Recall} & \textbf{F1-score} & \textbf{Precision} & \textbf{Recall} & \textbf{F1-score} & &  \\
        \hline
        ReLU  & ResNet-50 &  35.74 & 23.16 & 23.35 & 67.67 & 67.67 & 67.67 &  67.67 & 52.84\\
        LReLU & ResNet-50 & 35.97 & 23.56 & 23.65 & 68.10  & 68.10 & 68.10 &  68.10 & 53.37\\
        PReLU  & ResNet-50  & 36.30 & 23.89 & 23.77 & 68.77  & 68.77 & 68.77 & 68.77  & 54.10\\
        \textbf{ASAU}  & ResNet-50  & \textbf{46.29} & \textbf{31.30} & \textbf{34.15} &  \textbf{72.47} & \textbf{72.47} & \textbf{72.47} & \textbf{72.47}  & \textbf{61.10}\\
        \hline
        
        ReLU  & ResNet-18 & 33.70  & 25.83  & 27.43 & 70.12  & 70.12 & 70.12 & 70.12  & 57.10\\
        Leaky ReLU & ResNet-18 & 33.91 & 25.98 & 27.60 &  70.38 &  70.38 & 70.38 &  70.38 & 57.52\\
        PReLU  & ResNet-18 & 34.20 & 26.31 & 27.99 & 70.52  &  70.52 & 70.52 & 70.52  & 57.67\\
        \textbf{ASAU} & ResNet-18 & \textbf{36.50} & \textbf{27.30} & \textbf{28.99} &  \textbf{71.10} & \textbf{71.10} & \textbf{71.10} & \textbf{71.10}  & \textbf{58.59}\\
        
        \bottomrule
    \end{tabular}
    \label{CTscantable1}
\end{table*}

\begin{table*}[!t]
\scriptsize
    \centering
    \caption{ Activation functions and their impact across 26 Classes in RedImageNet  Abdominal/Pelvis MRI Scans.}
    \begin{tabular}{c|c|c|c|c|c|c|c|c|c}
       \toprule
        \multirow{2}{*}{\shortstack{\textbf{Activation}\\ \textbf{Function}}} & \multirow{2}{*}{\textbf{Method}} & \multicolumn{3}{c|}{\textbf{Macro Average}} & \multicolumn{3}{c|}{\textbf{Micro Average}} & \multirow{2}{*}{\textbf{Accuracy}}  & \multirow{2}{*}{\textbf{MCC}} \\
        \cline{3-8}
        & & \textbf{Precision} & \textbf{Recall} & \textbf{F1-score} & \textbf{Precision} & \textbf{Recall} & \textbf{F1-score} & &  \\
        \hline
        ReLU & ResNet-50 & 40.43 & 24.83 & 28.04 & 86.86  & 86.86 & 86.86 &  86.86 & 62.92\\
        LReLU & ResNet-50 & 40.56 & 24.99 & 28.56 &  87.10 & 87.10 & 87.10 &  87.10 & 63.25\\
        PReLU  & ResNet-50  & 41.10 & 25.20 & 28.81 & 87.25  & 87.25 & 87.25 & 87.25  & 63.47\\
        \textbf{ASAU} & ResNet-50  & \textbf{44.46} & \textbf{33.23} & \textbf{36.17} &  \textbf{89.20} & \textbf{89.20} & \textbf{89.20} &  \textbf{89.20} & \textbf{69.75}\\
        \hline
        
        ReLU  & ResNet-18 & 35.15 & 26.66 & 28.82 & 87.96  & 87.96  & 87.96 &   87.96 & 66.65\\
        LReLU & ResNet-18 & 35.67 &  26.87 & 28.99 & 88.12 & 88.12 & 88.12 & 88.12 &   66.72 \\
        PReLU  & ResNet-18 & 35.91 &  26.80 & 29.10 & 88.27 & 88.27 & 88.27 & 88.27 &   66.81\\
       \textbf{ASAU} & ResNet-18 & \textbf{38.58} & \textbf{27.58} & \textbf{29.67} & \textbf{88.62}  & \textbf{88.62} & \textbf{88.62} & \textbf{88.62}  & \textbf{67.32}\\
        
        \bottomrule
    \end{tabular}
    \label{MRIscantable2}
\end{table*}

\begin{table*}[!t]
\tiny
\centering
\caption{Comparison of different activation functions on liver segmentation benchmark (LiTS) dataset.}
\label{tab:segmentationlITS}
\begin{tabular}{c|c|c|c|c|c|c|c|c|c|c|c|c|c|c|c|c}
\toprule
\multirow{2}{*}{\shortstack{\textbf{Activation}\\ \textbf{Function}}} & \multicolumn{4}{c|}{\textbf{UNet~\cite{ronneberger2015u}}} & \multicolumn{4}{c|}{\textbf{DoubleUNet~\cite{jha2020doubleu}}} & \multicolumn{4}{c|}{\textbf{ColonSegNet~\cite{jha2021real}}} & \multicolumn{4}{c}{\textbf{UNext~\cite{unext}}} \\ \cline{2-17}
& \textbf{mDSC} & \textbf{mIoU} & \textbf{Rec.} & \textbf{Prec.} & \textbf{mDSC} & \textbf{mIoU} & \textbf{Rec.} &  \textbf{Prec.} &\textbf{mDSC} & \textbf{mIoU} & \textbf{Rec.} &  \textbf{Prec.} &\textbf{mDSC} & \textbf{mIoU} & \textbf{Rec.} &  \textbf{Prec.}  \\ \hline
ReLU    & 82.06  & 73.40 & 77.82 & 91.10 & 86.24 & 77.89 & 80.68 & 95.00 &80.87 &71.71  &\textbf{80.07 }& 87.50 & 80.27 & 71.31 & 78.84 & 87.82\\
LReLU  & 82.47 & 73.98 & 78.46 & 91.17 & 86.27 & 78.21 & 80.20 & 95.67 & 80.84 & 71.69  & 79.31 &  88.10 & 80.92 & 72.15 & 76.84 & \textbf{90.85} \\
PReLU    & 82.71 & 73.91 & 78.77 & 91.21 & 86.39 & 78.09 & 80.39 & 95.59 &78.91   & 69.85 &77.89  & 86.27 & 79.08 & 69.95 & 77.38 & 88.55\\ 

\textbf{ASAU} & \textbf{83.62}  &\textbf{75.08}  & \textbf{80.59} & \textbf{93.29}  & \textbf{86.88} & \textbf{78.73}  & \textbf{80.77} & \textbf{96.58} & \textbf{83.03} & \textbf{74.18} & 78.79 &  \textbf{91.17} & \textbf{81.98} & \textbf{73.03} & \textbf{79.01} & 90.45 \\ 
\bottomrule
\end{tabular}
\end{table*}

% \begin{figure*}
%     \centering
%      \includegraphics[width=0.4\textwidth]{isbi.drawio.png}
%     \caption{Liver segmentation results of the best performing DoubleUNet architecture.}
%     \label{fig:enter-label}
% \end{figure*}

\begin{table*}[!t]
\tiny
\centering
\caption{Comparison of different activation functions on liver tumor segmentation benchmark (LiTS) Benchmark dataset.}
\label{tab:segmentation2}
\begin{tabular}{c|c|c|c|c|c|c|c|c|c|c|c|c|c|c|c|c} %|c|c|c|c 
\toprule
\multirow{2}{*}{\shortstack{\textbf{Activation}\\ \textbf{Function}}} & \multicolumn{4}{c|}{\textbf{TransNetR~\cite{jha2023transnetr}}} & \multicolumn{4}{c|}{\textbf{TransResUNet~\cite{tomar2022transresu}}} & \multicolumn{4}{c|}{\textbf{ResUNet++~\cite{jha2019resunet++}}} & \multicolumn{4}{c}{\textbf{NanoNet-A~\cite{nanonet}}} \\ \cline{2-17}

& \textbf{mDSC} & \textbf{mIoU}  & \textbf{Rec. } & \textbf{Prec.}  &\textbf{mDSC} & \textbf{mIoU}  &\textbf{Rec.} & \textbf{Prec.} & \textbf{mDSC} & \textbf{mIoU}  &\textbf{Rec.} & \textbf{Prec.} & \textbf{mDSC} & \textbf{mIoU}  &\textbf{Rec.} & \textbf{Prec.}\\ \hline

ReLU &86.11  &77.95  &80.16  &96.34  & \textbf{86.38} & \textbf{78.23}  & \textbf{ 80.85}   &95.77 & 77.03 & 68.19 & 79.90 & 83.57 & 75.05 & 66.53 & 73.33 & 83.25  \\

LReLU  & 86.30 & 78.37   & \textbf{80.37} & 96.67 &86.18  & 77.82  &79.80  & 96.59 & 75.03 & 66.35 & 71.64 & \textbf{84.05} & 74.05 & 65.02 & 73.36 & 84.23 \\

PReLU  &  86.39 & 78.29 & 80.32 & 96.52   &85.68 &77.30 &  79.46 & 96.12 & 74.39  & 66.02 & 74.38 & 81.99 & 74.84 & 66.08 & 72.47 & 85.24\\

\textbf{ASAU} & \textbf{86.41} & \textbf{78.49} &  80.30 & \textbf{96.65} & 86.35 & 78.15  & 79.89  & \textbf{96.67} & \textbf{78.12} & \textbf{69.56} &  \textbf{81.50} & 83.81 & \textbf{76.57} & \textbf{67.30} & \textbf{75.88} & \textbf{84.61}\\ 
\bottomrule
\end{tabular}
\end{table*}

\subsection{Datasets}
For the classification tasks, we use RadImageNet~\cite{mei2022radimagenet} database, which is an open-access medical imaging database designed to improve transfer learning performance on downstream medical imaging applications and perhaps the largest ever medical imaging dataset so far. From the whole dataset, we experiment on CT abdominal/pelvis, consisting of 28 disease classes with an average class size of 4994 and a total of 139,825 slices (i.e., the dataset is designed to have slices per disease although the scans are volumes-3D). The CT dataset contains 28 (disease) classes such as adrenal pathology, arterial pathology, ascites, bil dil, bladder pathology, bowel abnormality, bowel inflammation, bowel mass degenerative changes, dilated urinary tract, fat-containing tumor, gallbladder pathology, gallstone, intraperitoneal mass, liver lesion, normal osseous, neoplasm, ovarian pathology, pancreatic lesion, post-op, prostate lesion, renal lesion, soft tissue collection, soft tissue mass, splenic lesion, urolithiasis and uterine pathology. We also experimented on the MRI abdomen/pelvis dataset consisting of 26 distinct (disease) classes with an average class size of 3513 slices and a total number of 91,348 slices. While most of these classes overlap with the CT dataset, the MRI dataset uniquely includes classes like enlarged organs and liver disease, which are not in the CT dataset. The CT dataset, conversely, has a specific class for entire abdominal organs.

%While most of these classes were present in the MRI dataset except for enlarged organs, liver disease, and post-op whereas, CT has a different abdominal entire organ class. 

For the segmentation tasks, we select the Liver Tumor Segmentation Benchmark (LiTS)~\cite{bilic2023liver} dataset, which is a multi-center dataset collected from seven clinical centers. It contains 201 CT images of the abdomen. The dataset is completely anonymized, and the images have been reviewed visually to preclude the presence of personal identifiers. The whole dataset is distributed into a training dataset with 130 CT scans, and the test dataset has 71 CT scans. Only the training dataset is made publically available. Thus, we trained the dataset for our experimentation. 

%\vspace{-7mm}

\subsection{Implementation details}
We used the PyTorch~\cite{paszke2019pytorch} framework for all the segmentation experiments. The networks were configured to train for liver segmentation tasks with a batch size of 16 and a learning rate set to 1e$^{-4}$. 500 epochs of training were performed to fine-tune the network parameters adequately with an early stopping patience of 50. To enhance the performance of our network,  we used a combination of binary cross-entropy and dice loss, and an Adam optimizer was chosen for parameter updates. The data was split into 80\% for training, 10\% for validation, and 10\% for testing. We resized the image to $256\times256$ pixels in-plane resolution to optimize the trade-off between training time and model complexity. To avoid bias, we also split the cases into independent training (70 patients), validation (30 patients), and test (30 patients) sets. The volumetric CT scans were processed pseudo-3D (slice-by-slice) to fit into regular computer hardware (GPU). During prepossessing, we extracted the healthy liver masks. 

For classification (disease detection) experiments, we consider Tensorflow-Keras~\cite{keras} framework. We consider ResNet-18~\cite{resnet} and ResNet-50~\cite{resnet} as our baseline classification networks. The networks are trained with batch size 32, initial learning rate 0.00001 with Adam~\cite{adam} optimizer and $1e^{-4}$ weight decay rate. The data was split into 80\% for training, 10\% for validation, and 10\% for testing. Results are reported on CT scan image data in Table~\ref{CTscantable1} and MRI image data in Table~\ref{MRIscantable2}.

%Thus, we have 11684 slices in training dataset, 2745 in validation, and 4734 slices in the testing set. The difference in slices in the validation and test is because the number of slices might vary between different patients. 

%In our experiments, we designed a network adhering to the PyTorch framework, exploiting the computational power of an Nvidia-RTX 2080TI GPU. The network was configured to train with a batch size of 1, and a learning rate set to 0.0001. A total of 100 epochs of training were performed to fine-tune the network parameters adequately. This training regimen was meticulously chosen to balance the trade-off between computational efficiency and the convergence of model accuracy. Our implementation paid special attention to the memory constraints and computational load by integrating a lightweight encoder structure that employs the innovative Ghost module, aiming to reduce the number of network parameters without compromising the model's segmentation accuracy.

\subsection{Results and Discussion}
Table~\ref{CTscantable1} shows the results of ResNet-50~\cite{he2016deep} and ResNet-18 architectures on the  CT abdominal/pelvis scan dataset. Here, we examine the efficacy of ReLU, LReLU, and PReLU activation along with our proposed ASAU. For all the experiments, the ASAU-based method outperformed all other experimental settings in all the metrics with a significant margin. On the ResNet-50-based architecture, the ASAU integrated method obtains a high improvement of 8.26\% in terms of MCC. Table~\ref{MRIscantable2} shows the MRI abdominal/pelvis dataset results. Again, here, ASU-integrated ResNet50 obtained an MCC score of 69.75\%, which is 6.83\% better than ReLU based method. 

Table~\ref{tab:segmentationlITS} and Table~\ref{tab:segmentation2} show LiTs datasets' results. Here, we have compared the performance of eight state-of-the-art medical image segmentation methods (UNet~\cite{ronneberger2015u}, DoubleUNet~\cite{jha2020doubleu}, ColonSegNet~\cite{jha2021real}, UNext~\cite{unext}, TransNetR~\cite{jha2023transnetr}, TransResNet~\cite{tomar2022transresu}, ResUNet++~\cite{jha2019resunet++} and NanoNet-A~\cite{nanonet}) with four different activation functions. From the table, it can be observed that ASAU has significant improvement from 1\% to 3\% as compared to the ReLU activation function. DoubleUNet, along with ASAU, has set a new baseline for liver segmentation tasks with a high dice score of 86.88\%, mIoU of 78.73\%, recall of 80.77\%, and precision of 96.58\%.

% \subsection{Discussion}
% The robustness of SmoothDoubleUNet to variations in image quality and the presence of noise, hallmarks of real-world clinical data, underscores its potential for revolutionizing liver segmentation practices. Incorporating the smooth activation function has been pivotal in enhancing model performance, particularly in low-contrast scenarios that traditionally challenge segmentation algorithms. Such a leap in accuracy and reliability can translate into significant clinical advancements, from diagnostic accuracy to surgical planning. While our model specializes in binary segmentation, its adaptability and foundational principles present clear pathways for future exploration in multi-organ segmentation tasks.

\section{Conclusion}
\label{conclusion}

% In this paper, we present the Adaptive Smooth activation function, an innovative active function for medical image classification and segmentation that, when integrated  into CNN enhances classification and segmentation accuracy. Our approach leverages variable perturbations during training, significantly increasing the model's robustness to variations in liver shape and pathology. In comparative analyses, SmoothDoubleUNet demonstrates marked improvements in performance over standard U-Net models, especially in challenging scenarios characterized by noise and blurred edges in CT images. Our experiments reveal that the fine-tuning of the network, particularly the image encoders while keeping the mask decoders fixed results in superior segmentation outcomes. While the current iteration of SmoothDoubleUNet is tailored for binary segmentation of liver tissues, future developments will expand its capabilities to multiclass segmentation and address the segmentation of multiple organs. Despite these forthcoming enhancements, SmoothDoubleUNet offers a promising new tool for medical centers, simplifying the integration of advanced segmentation technology with minimal annotation requirements.

Our study introduces a pioneering activation function, Adaptive Smooth Maximum Unit (ASAU), to enhance the classification of abdominal organs and segment abdominal liver CT images. The promising results attained by our framework on the RadImageNet and  LiTS datasets underscore its potential applicability to a broader range of organ imaging tasks. Future research endeavors are directed towards harnessing three-dimensional spatial information more effectively, aiming to refine the diagnostic tools available to clinicians further and potentially extend the application of our method to include a comprehensive suite of abdominal organs, enhancing our approach's overall robustness and generalization.

%Our findings demonstrate a commendable improvement in classification accuracy and segmentation precision supported by a notable reduction in network parameters—a testament to the ASAU's capability of addressing the complexity of medical image analysis while maintaining computational efficiency.

%This structured presentation of your dataset, implementation details, and conclusions offers a clear and concise overview of the essential aspects of your research paper, setting the stage for more in-depth discussions in the subsequent sections.
\vspace{-3mm}
\subsection*{Compliance with Ethical Standards}
This research study was conducted retrospectively using human subject data made available in open access by. Ethical approval was not required, as confirmed by the license attached with the open-access data.

\vspace{-3mm}
\subsection*{Conflicts of Interest}
 The authors have no relevant financial interests to disclose. %or non-financial

\bibliographystyle{unsrtnat}
\bibliography{references}  %%% Uncomment this line and comment out the ``thebibliography'' section below to use the external .bib file (using bibtex) .

\begin{thebibliography}{23}
\providecommand{\natexlab}[1]{#1}
\providecommand{\url}[1]{\texttt{#1}}
\expandafter\ifx\csname urlstyle\endcsname\relax
  \providecommand{\doi}[1]{doi: #1}\else
  \providecommand{\doi}{doi: \begingroup \urlstyle{rm}\Url}\fi

\bibitem[Reyes et~al.(2008)Reyes, Ballester, Li, Kozic, Summers, and Linguraru]{reyes2008interpretability}
Mauricio Reyes, Miguel A~Gonzalez Ballester, Zhixi Li, Nina Kozic, Ronald~M Summers, and Marius~George Linguraru.
\newblock Interpretability of anatomical variability analysis of abdominal organs via clusterization of decomposition modes.
\newblock In \emph{Proceedings of the International Conference of the IEEE Engineering in Medicine and Biology Society}, pages 355--358, 2008.

\bibitem[Arnold et~al.(2020)Arnold, Abnet, Neale, Vignat, Giovannucci, McGlynn, and Bray]{arnold2020global}
Melina Arnold, Christian~C Abnet, Rachel~E Neale, Jerome Vignat, Edward~L Giovannucci, Katherine~A McGlynn, and Freddie Bray.
\newblock Global burden of 5 major types of gastrointestinal cancer.
\newblock \emph{Gastroenterology}, 159\penalty0 (1):\penalty0 335--349, 2020.

\bibitem[Jha et~al.(2020)Jha, Riegler, Johansen, Halvorsen, and Johansen]{jha2020doubleu}
Debesh Jha, Michael~A Riegler, Dag Johansen, P{\aa}l Halvorsen, and H{\aa}vard~D Johansen.
\newblock {DoubleU-Net: A Deep Convolutional Neural Network for Medical Image Segmentation}.
\newblock In \emph{Proceedings of the IEEE 33rd International Symposium on computer-based medical systems (CBMS)}, pages 558--564, 2020.

\bibitem[Tomar et~al.(2022)Tomar, Shergill, Rieders, Bagci, and Jha]{tomar2022transresu}
Nikhil~Kumar Tomar, Annie Shergill, Brandon Rieders, Ulas Bagci, and Debesh Jha.
\newblock {TransResU-Net:} {Transformer based ResU-Net for real-time colonoscopy polyp segmentation}.
\newblock \emph{arXiv preprint arXiv:2206.08985}, 2022.

\bibitem[jha(2023)]{jha2023transnetr}
\emph{{TransNetR:} {Transformer-based Residual Network for Polyp Segmentation with Multi-Center Out-of-Distribution Testing}}, 2023.

\bibitem[Nair and Hinton(2010)]{relu}
Vinod Nair and Geoffrey~E Hinton.
\newblock Rectified linear units improve restricted boltzmann machines.
\newblock In \emph{Proceedings of the 27th international conference on machine learning (ICML-10)}, pages 807--814, 2010.

\bibitem[Xu et~al.(2015)Xu, Wang, Chen, and Li]{xu2015empirical}
Bing Xu, Naiyan Wang, Tianqi Chen, and Mu~Li.
\newblock Empirical evaluation of rectified activations in convolutional network.
\newblock \emph{arXiv preprint:1505.00853}, 2015.

\bibitem[He et~al.(2015)He, Zhang, Ren, and Sun]{he2015delving}
Kaiming He, Xiangyu Zhang, Shaoqing Ren, and Jian Sun.
\newblock Delving deep into rectifiers: Surpassing human-level performance on imagenet classification.
\newblock In \emph{Proceedings of the IEEE international conference on computer vision}, pages 1026--1034, 2015.

\bibitem[Maas et~al.(2013)Maas, Hannun, Ng, et~al.]{lrelu}
Andrew~L Maas, Awni~Y Hannun, Andrew~Y Ng, et~al.
\newblock Rectifier nonlinearities improve neural network acoustic models.
\newblock In \emph{Proc. icml}, volume~30, page~3, 2013.

\bibitem[Misra(2020)]{mish}
Diganta Misra.
\newblock Mish: A self regularized non-monotonic activation function, 2020.

\bibitem[Goodfellow et~al.(2013)Goodfellow, Warde-Farley, Mirza, Courville, and Bengio]{maxout}
Ian~J. Goodfellow, David Warde-Farley, Mehdi Mirza, Aaron Courville, and Yoshua Bengio.
\newblock Maxout networks, 2013.

\bibitem[Ronneberger et~al.(2015)Ronneberger, Fischer, and Brox]{ronneberger2015u}
Olaf Ronneberger, Philipp Fischer, and Thomas Brox.
\newblock {U-Net}: {Convolutional Networks for Biomedical Image Segmentation}.
\newblock In \emph{Proceedings of the 18th International Conference on Medical Image Computing and Computer-Assisted Intervention}, pages 234--241, 2015.

\bibitem[Jha et~al.(2021{\natexlab{a}})Jha, Ali, Tomar, Johansen, Johansen, Rittscher, Riegler, and Halvorsen]{jha2021real}
Debesh Jha, Sharib Ali, Nikhil~Kumar Tomar, H{\aa}vard~D Johansen, Dag Johansen, Jens Rittscher, Michael~A Riegler, and P{\aa}l Halvorsen.
\newblock {Real-Time Polyp Detection, Localization and Segmentation in Colonoscopy using Deep Learning}.
\newblock \emph{IEEE Access}, 9:\penalty0 40496--40510, 2021{\natexlab{a}}.

\bibitem[Valanarasu and Patel(2022)]{unext}
Jeya Maria~Jose Valanarasu and Vishal~M. Patel.
\newblock Unext: Mlp-based rapid medical image segmentation network, 2022.

\bibitem[Jha et~al.(2019)Jha, Smedsrud, Riegler, Johansen, De~Lange, Halvorsen, and Johansen]{jha2019resunet++}
Debesh Jha, Pia~H Smedsrud, Michael~A Riegler, Dag Johansen, Thomas De~Lange, P{\aa}l Halvorsen, and H{\aa}vard~D Johansen.
\newblock {ResUNet++:} {An Advanced Architecture for Medical Image Segmentation}.
\newblock In \emph{Proc. of int. symposium on multimedia}, pages 225--2255, 2019.

\bibitem[Jha et~al.(2021{\natexlab{b}})Jha, Tomar, Ali, Riegler, Johansen, Johansen, de~Lange, and Halvorsen]{nanonet}
Debesh Jha, Nikhil~Kumar Tomar, Sharib Ali, Michael~A Riegler, H{\aa}vard~D Johansen, Dag Johansen, Thomas de~Lange, and P{\aa}l Halvorsen.
\newblock Nanonet: Real-time polyp segmentation in video capsule endoscopy and colonoscopy.
\newblock In \emph{Proceedings of the International Symposium on Computer-Based Medical Systems (CBMS)}, pages 37--43, 2021{\natexlab{b}}.

\bibitem[Mei et~al.(2022)Mei, Liu, Robson, Marinelli, Huang, Doshi, Jacobi, Cao, Link, Yang, et~al.]{mei2022radimagenet}
Xueyan Mei, Zelong Liu, Philip~M Robson, Brett Marinelli, Mingqian Huang, Amish Doshi, Adam Jacobi, Chendi Cao, Katherine~E Link, Thomas Yang, et~al.
\newblock {RadImageNet}: {RadImageNet: An Open Radiologic Deep Learning Research Dataset for Effective Transfer Learning}.
\newblock \emph{Radiology: Artificial Intelligence}, 4\penalty0 (5):\penalty0 e210315, 2022.

\bibitem[Bilic et~al.(2023)Bilic, Christ, Li, Vorontsov, Ben-Cohen, Kaissis, Szeskin, Jacobs, Mamani, Chartrand, et~al.]{bilic2023liver}
Patrick Bilic, Patrick Christ, Hongwei~Bran Li, Eugene Vorontsov, Avi Ben-Cohen, Georgios Kaissis, Adi Szeskin, Colin Jacobs, Gabriel Efrain~Humpire Mamani, Gabriel Chartrand, et~al.
\newblock The liver tumor segmentation benchmark (lits).
\newblock \emph{Medical Image Analysis}, 84:\penalty0 102680, 2023.

\bibitem[Paszke et~al.(2019)Paszke, Gross, Massa, Lerer, Bradbury, Chanan, Killeen, Lin, Gimelshein, Antiga, et~al.]{paszke2019pytorch}
Adam Paszke, Sam Gross, Francisco Massa, Adam Lerer, James Bradbury, Gregory Chanan, Trevor Killeen, Zeming Lin, Natalia Gimelshein, Luca Antiga, et~al.
\newblock Pytorch: An imperative style, high-performance deep learning library.
\newblock \emph{Advances in neural information processing systems}, 32, 2019.

\bibitem[Chollet et~al.(2015)]{keras}
Fran\c{c}ois Chollet et~al.
\newblock Keras.
\newblock \url{https://keras.io}, 2015.

\bibitem[He et~al.(2016{\natexlab{a}})He, Zhang, Ren, and Sun]{resnet}
Kaiming He, Xiangyu Zhang, Shaoqing Ren, and Jian Sun.
\newblock Deep residual learning for image recognition.
\newblock In \emph{Proceedings of the IEEE conference on computer vision and pattern recognition}, pages 770--778, 2016{\natexlab{a}}.

\bibitem[Kingma and Ba(2017)]{adam}
Diederik~P. Kingma and Jimmy Ba.
\newblock Adam: A method for stochastic optimization, 2017.

\bibitem[He et~al.(2016{\natexlab{b}})He, Zhang, Ren, and Sun]{he2016deep}
Kaiming He, Xiangyu Zhang, Shaoqing Ren, and Jian Sun.
\newblock Deep residual learning for image recognition.
\newblock In \emph{Proceedings of the IEEE conference on computer vision and pattern recognition}, pages 770--778, 2016{\natexlab{b}}.

\end{thebibliography}





% \end{thebibliography}

\end{document}